\documentclass[fleqn,10pt,twocolumn]{AROB}
\usepackage{amsmath}
\usepackage{amsfonts}
\usepackage{mathtools}
\newcommand{\argmax}{\mathop{\rm arg~max}\limits}

\title{Neural Risk-sensitive Satisficing in Contextual Bandits}

\author{Shogo Ito${}^{1\dagger}$, Tatsuji Takahashi${}^{1}$ and Yu Kono${}^{1}$}
\speaker{Shogo Ito}

\affils{${}^{1}$School of Science and Engineering, Tokyo Denki University, Saitama, Japan\\
(Tel: +81-49-296-5416; E-mail: yu.kohno.02@gmail.com)\\
}

\abstract{%
The contextual bandit problem, which is a type of reinforcement learning tasks, provides an effective framework for solving challenges in recommendation systems, such as satisfying real-time requirements, enabling personalization, addressing cold-start problems.
However, contextual bandit algorithms face challenges since they need to handle large state-action spaces sequentially.
These challenges include the high costs for learning and balancing exploration and exploitation, as well as large variations in performance that depend on the domain of application.
To address these challenges, Tsuboya et~al. proposed the Regional Linear Risk-sensitive Satisficing ($\mathrm{RegLinRS}$) algorithm. $\mathrm{RegLinRS}$ switches between exploration and exploitation based on how well the agent has achieved the target.
However, the reward expectations in $\mathrm{RegLinRS}$ are linearly approximated based on features, which limits its applicability when the relationship between features and reward expectations is non-linear.
To handle more complex environments, we proposed Neural Risk-sensitive Satisficing ($\mathrm{NeuralRS}$), which incorporates neural networks into $\mathrm{RegLinRS}$, and demonstrated its utility.
}


\begin{document}

\maketitle


\section{Introduction}
%
Recommendation and advertisement delivery systems are indispensable components of modern web services.
%
To maximize their effectiveness, machine learning models need to make data-driven decisions.
%
These models typically rely on historical data to learn the user's interests and preferences.
%
Recently, counterfactual machine learning approaches have been proposed, which allow the agent to perform offline evaluation of action data that it has not previously experienced.
%
However, these approaches alone are not sufficient. Real-time (online) data sampling and feedback are necessary to personalize recommendation systems for individual users, address the cold-start problem, and adapt to changes in environment.
%
Contextual bandit algorithms, which can reflect user preferences to optimize recommendation choices, are good at making sequential decisions and can adapt to these challenges.
%
Considering these factors, we expect future recommendation systems would require hybrid approaches that combine offline and online methods.
%
However, contextual bandit algorithms, which sequentially handle vast state-action spaces, face challenges such as high costs in learning and in exploration-and-exploitation. Their performance significantly varies depending on different application domains, and that limits their guarantees of optimization.
%
In these aspects, the contextual bandit algorithms still face many challenges. 
%
Even so, in industrial applications, developers tend to prioritize reaching the target over optimization, since the latter requires extensive exploration.
%
Moreover, online reinforcement learning is a practical approach for real-world environments, where data and computional resources are limited. It enables sequential learning of optimal decision-making and action selection based on the outcomes.
%
We focused on an approach that prioritizes reaching a target over long-term optimization.
%
%
As a prior study of such an approach, Tsuboya et~al. proposed the Regional Linear Risk-sensitive Satisficing ($\mathrm{RegLinRS}$) algorithm, which extended the basic Risk-sensitive Satisficing ($\mathrm{RS}$) algorithm \cite{takahashi16}. 
%
$\mathrm{RS}$ switches between exploration and exploitation based on target achievement status.
%
Tsuboya et~al. incorporated linear approximation for state estimation to extend $\mathrm{RS}$ into $\mathrm{RegLinRS}$ \cite{tsuboya24}.
%
They showed that $\mathrm{RegLinRS}$ outperforms benchmark algorithms in contextual bandits.
%
$\mathrm{RegLinRS}$ performs well, at least, in environments where the relationship between features and the expected rewards computed by the model, is linear.
%
By extending $\mathrm{RegLinRS}$ to handle non-linear relationships, we expect the algorithm become capable of addressing a wider range of problems and achieve greater performance improvements compared to its linear version.
%
In this paper, we propose Neural Risk-sensitive Satisficing ($\mathrm{NeuralRS}$), an algorithm that leverages Neural Network (NN) as function approximators to extend $\mathrm{RegLinRS}$.
%
This extension enables $\mathrm{NeuralRS}$ to handle non-linear relationships between features and expected rewards, as well as high-dimensional data.
%
Our objective is to demonstrate that the $\mathrm{NeuralRS}$ algorithm can adapt to environments more flexibly than $\mathrm{RegLinRS}$ in a real-world dataset.
%
We also aim to show that $\mathrm{NeuralRS}$ achieves comparable or higher performance than that of typical NN-based algorithms.

\section{Problem Setting}
%
The contextual bandit problem is a framework that abstracts the task of maximizing rewards by estimating the expected rewards of actions using feature-based information.
%
At each time step $t$, the agent is provided with a $d$-dimensional feature vector ${\bf x}_t$.
%
The agent chooses an action based on this feature vector and receives a reward $r_{t}$.
%
The reward $r_t$ at time $t$ is defined as shown in Eq. \eqref{eq:reward}.
\begin{align}
    \label{eq:reward}
    r_t = h({\bf x}_{t}) + \epsilon _t
\end{align}
%
$h({\bf x}_{t})$ is unknown reward function and $\epsilon _t$ is an error term whose expectation is $0$ at time $t$.
%
When the relationship between the feature vector ${\bf x}_{t}$ and the expected reward is linear, $h({\bf x}_{t})$ can be expressed as $h({\bf x}_{t}) = {\boldsymbol \theta}^{\top} {\bf x}_{t}$, 
%
where ${\boldsymbol \theta}$ is an unknown parameter vector associated with the selected action.
%
On the other hand, when the relationship is non-linear, neural bandit algorithms such as $\mathrm{NeuralUCB}$ \cite{zhou20} and $\mathrm{NeuralTS}$ \cite{zhang21} use neural networks to approximate the relationship.
%
In this case, $h({\bf x}_t)$ is defined as the output of the neural network, enabling the algorithm to represent complex functions.
%
Contextual bandit algorithms learn the parameter ${\boldsymbol \theta}$ to estimate the action value for the feature vector.
%
To evaluate performance, we use the metric of regret.
%
Regret shows how much difference or loss there is between the ideal case where the agent always took the best action and the actual case observed as the action of the agent.
%
A lower regret value indicates better performance of the agent in maximizing rewards.
%
The method for calculating regret is shown in Eq. (\ref{eq:regret}).
\hspace{-4pt}
\begin{align}
    \label{eq:regret}
    \mathrm {Regret}{(T)} &= \sum^T_{t=1}(p_{t,i^\ast} - p_{t,i}) \geq 0
\end{align}
\hspace{-4pt}
%
Here, $p_{t,i^\ast}$ represents the highest expected reward among the given actions at time $t$, and $p_{t,i}$ is the expected reward of the action selected at time $t$.
%
Well-known algorithms that demonstrate good performance in contextual bandit problem include $\mathrm{LinUCB}$ \cite{li10} and $\mathrm{LinTS}$ \cite{riquelme18}.
%
However, when adaptation to non-linear models is required, algorithms such as $\mathrm{NeuralUCB}$ and $\mathrm{NeuralTS}$ are also effective methods.

\section{Risk-sensitive Satisficing}
%
Takahashi et~al. proposed the Risk-sensitive Satisficing ($\mathrm{RS}$) algorithm \cite{takahashi16} to model satisficing \cite{simon56}.
%
$\mathrm{RS}$ aims to discover actions that exceed an aspiration level and switches to exploitation, stopping more exploration as soon as one of such actions is found.
%

\subsection{Definition of subjective regret}
%
The standard regret, as defined in Eq. \eqref{eq:regret}, is an objective measure that quantifies  the difference between the actual cumulative reward and the ideal cumulative reward, which assumes consitently taking the optimal action from the start.
%
The agent aims to minimize this objective metric, although it cannot directly observe the value.
%
In contrast, subjective regret ($\mathrm{SR}$) is the cumulative difference between $\aleph$ (aleph) and $r$, representing how much the rewards fall short of the aspiration level.
%
Thus, $\mathrm{SR}$ can be utilized as an internal measure for decision-making, as it is directly observable by the agent.
%
Eq. \eqref{eq:subjective_regret} defines $\mathrm{SR}$ for each action.
\begin{align}
    \label{eq:subjective_regret}
    I^{\mathrm{SR}}_{i} = \sum_{t=1}^N (\aleph - p_{t, \mathrm{select}})
\end{align}

\subsection{Definition of RS}
%
Assuming that $I^{\mathrm{RS}}_i \coloneqq - I^{\mathrm{SR}}_i$ represents the subjective value of action $i$ after the agent has chosen it $n$ times, the value function $I^{\mathrm{SR}}_i$ for $\mathrm{RS}$ is defined by Eq. \eqref{eq:RS}.
%
At each time step, the agent selects the action that maximizes the expected value $I^{\mathrm{RS}}_{i}$.
\begin{align}
    \label{eq:RS}
    I^{\mathrm{RS}}_{i} &= \frac{n_i}{N}(E_i - \aleph)
\end{align}
%
where $n_i$ represents the number of times the agent has chosen action $a_i$, $N$ represents the total number of trials, $E_i$ represents the expected reward obtained from action $a_i$, and $\aleph$ represents the aspiration level.
%
$\mathrm{RS}$ adopts the trial ratio $n_i / N$ as the reliability for $E_i$. 
%
$\mathrm{RS}$ promotes optimistic exploration for actions with a low trial ratio $n_i / N$ in cases where $E_i < \aleph$ (under-achieved).
%
Conversely, it promotes pessimistic exploitation for actions with a high trial ratio $n_i / N$ in cases where $E_i \geq \aleph$ (over-achieved).
%
$\mathrm{RS}$ has been proven to keep regret finite in the K-armed bandit problems \cite{tamatsukuri19}.

\section{Regional Linear RS}
%
Regional Linear Risk-sensitive Satisficing ($\mathrm{RegLinRS}$) was proposed as a function approximation method for $\mathrm{RS}$ that take into account features \cite{tsuboya24}.
%
$\mathrm{RegLinRS}$ have an episodic memory that stores feature vectors obtained in the past and concatenates them with action selection records.
%
It approximates and estimates the reliability $n_i / N$ and the average reward $E_i$, which are used in $\mathrm{RS}$, by leveraging episodic memory and linear regression, then applies these estimates to the value function.
%
The value function of $\mathrm{RegLinRS}$ is defined in the same structure as Eq. (\ref{eq:RS}) using approximated trial ratio ${\phi}_i$, the unbiased estimator ${\boldsymbol \theta}^{\top}_i {\bf x}_{t}$ for the feature vector ${\bf x}_{t}$, and the aspiration level $\aleph$, as shown in Eq. \eqref{eq:RegLinRS}.
\begin{align}
\label{eq:RegLinRS}
    I^{\mathrm{RegLinRS}}_{i} &= {\phi}_{i} \left({\boldsymbol \theta}^{\top}_i {\bf x}_{t}-\aleph\right)
\end{align}
%
At each time step, the agent selects the action that maximizes the expected value $I^{\mathrm{RegLinRS}}_{i}$.

\subsection{Linear approximation of expected rewards}
%
The expected rewards in $\mathrm{RegLinRS}$ are approximated using linear regression with the least squares method.
%
When action $a$ is selected at time $t$, the loss function between the received reward $r_t$ and the unbiased estimate $\boldsymbol{\theta}^{\top}_i {\mathbf{x}}_t$ is minimized using least squares, as defined in Eq. \eqref{theta_hat}.
\begin{align}
    \label{theta_hat}
    {\boldsymbol \theta^{\prime}} &= {\{\mathbf{I}} + \sum^{T}_{t=1} {\mathrm{\bf x}}_{t} {\mathrm{\bf x}}^{\top}_{t}\}^{-1} {\sum^T_{t=1} r_t {\mathrm{\bf x}}_{t}}
\end{align}

\subsection{Local approximation of the reliability} 
%
$\mathrm{RegLinRS}$ approximates the trial ratio for each action at the current state by leveraging feature vectors from similar states stored in episodic memory.
%
$\mathrm{RegLinRS}$ extracts $k$ samples (feature vectors and action selection records) from the episodic memory that are most similar (i.e., closest in distance) to the current feature vector ${\mathbf{x}}_t$.
%
These samples include feature vectors and action selection records, and the trial ratio approximation $\boldsymbol{\phi}$ is defined as in Eq. \eqref{phi}.
\begin{align}
    {\mathrm{Sim}}_{{\bf x}_{t},{\bf x}_{j}} &= \epsilon \left[ \frac{d^2_{{\bf x}_t,{{\bf x}_j}}}{\Bar{d^2}} + \epsilon \right]^{-1} \in (0, 1] \notag \\
    {w}_{j} &= \frac{{\mathrm{Sim}}_{{\bf x}_{t},{\bf x}_{j}}}{\sum^k_{j=1}{\mathrm{Sim}}_{{\bf x}_{t},{\bf x}_{j}}} \notag \\
    \label{phi}
    {\boldsymbol \phi} &= \sum^k_{j=1}{w}_{j}{\bf u}_{j}
\end{align}
%
$d^2_{{\bf x}_t,{{\bf x}}_j}$ represents the squared Euclidean distance, $\Bar{d^2}$ is the average of $d^2_{{\bf x}_t,{{\bf x}}_j}$, and $\epsilon$ is a small constant to avoid division by zero. 
%
Using these components, the agent calculates the expected value $I^{\mathrm{RegLinRS}}_{i}$ in Eq. \eqref{eq:RegLinRS}.

\section{Limitations of RegLinRS and extentions}
%
In real-world datasets, it is often unclear whether the relationship between features and expected rewards at each time step is linear or non-linear.
%
Tsuboya et~al. have shown that $\mathrm{RegLinRS}$ achieves the best performance compared to the benchmark algorithms in environments where the relationship is linear \cite{tsuboya24}.
%
However, the performance of $\mathrm{RegLinRS}$ in non-linear environments remains insufficiently explored.
%
Since $\mathrm{RegLinRS}$ uses linear regression for estimating expected rewards, it will most likely faces difficulties in handling complex state representations.
%
If we can make a new model by extending $\mathrm{RegLinRS}$ with Neural Networks, then the new model can address a wider range of problems compared to the linear model.

\section{Neural RS}
%
We propose Neural Risk-sensitive Satisficing ($\mathrm{NeuralRS}$), an algorithm that extends $\mathrm{RegLinRS}$ by incorporating Neural Networks (NN) as function approximators.
%
$\mathrm{NeuralRS}$ learns action values based on feature vectors provided at each time step using Mean Squared Error (MSE).
%
It also estimates trial ratios using k-means centroid updates.
%
Then, it applies the estimated expected rewards and the approximated reliability to its value function.
%
The value function of $\mathrm{NeuralRS}$ is defined as Eq. \eqref{eq:NeuralRS}, structured similarly to Eq. \eqref{eq:RS}.
\begin{align}
\label{eq:NeuralRS}
    I^{\mathrm{NeuralRS}}_{i} &= {\rho}_i \left(f_i({\bf x}_{t} ; {\boldsymbol \theta}_{t-1}) - \aleph \right)
\end{align}
%
${\rho}_i$ represents the approximate estimator of the trial ratio, $f_i({\bf x}_{t} ; {\boldsymbol \theta}_{t-1})$ represents the expected reward when action $a_i$ is selected based on the given feature vector ${\bf x}_{t}$, and $\aleph$ represents the aspiration level.
%
At each time step, the agent selects the action that maximizes the expected value $I^{\mathrm{NeuralRS}}_{i}$.
\begin{align}
    a^{\mathrm{select}} &= \argmax_i \Bigl(I^{\mathrm{NeuralRS}}_{i}\Bigr) \notag
\end{align}

\subsection{Non-linear approximation of expected rewards}
%
The NN used in $\mathrm{NeuralRS}$ takes a $d$-dimensional feature vector ${\bf x}_t$ as input at each time step, applies non-linear transformations through hidden layers, and outputs the expected reward $f_i({\bf x}_{t} ; {\boldsymbol \theta}_{t-1})$ for each action in the output layer.
%
In $\mathrm{NeuralRS}$, the expected reward $f({\bf x}_t ; {\boldsymbol{\theta}}_{t-1})$ is calculated as shown in Eq. \eqref{f}.
\begin{align}
    f^{(1)} &= {\mathrm{W}}^{(1)} {\bf x}_t \notag \\
    f^{(l)} &= {\mathrm{W}^{(l)}} {\mathrm{ReLU}}{(f^{(l-1)})} \quad 2 \leq l \leq L \notag \\
    \label{f}
    f({\bf x}_t; {\boldsymbol{\theta}}_{t-1}) &= \sqrt{g} f^{(l)} 
\end{align}
%
Here, $\mathrm{ReLU}$ is defined as $\max\{x, 0\}$, $g$ represents the width of the NN, and $\mathrm{W}^{(1)} \in {\mathbb{R}}^{ \times d}$, $\mathrm{W}^{(l)} \in {\mathbb{R}}^{g \times g}$, $2 \leq l < L$, and $\mathrm{W}^{(L)} \in {\mathbb{R}}^{1 \times g}$ are weight matrices.
%
In $\mathrm{NeuralRS}$, $f_i({\bf x}_t ; {\boldsymbol{\theta}}_{t-1})$ serves as the estimated expeceted reward for action $a_i$.
%
The parameter ${\boldsymbol{\theta}}$ is trained through minimizing the loss function $L({\boldsymbol{\theta}})$, defined in Eq. \eqref{eq:loss}, which is based on the Mean Squared Error (MSE).
\begin{align}
    \label{eq:loss}
    \min_{\theta} L({\boldsymbol{\theta}}) &= \frac{1}{2}\sum_{t=1}^{T} \left[f_i({\bf x}_t; {\boldsymbol{\theta}}_{t-1}) - r_{t}\right]^2
\end{align}
%
Here, $r_t$ represents the actual reward obtained when action $a_i$ is selected at time $t$.

\subsection{Local approximation of the reliability} 
\begin{figure}[hbtp]
    \begin{center}
        \includegraphics[width=8cm]{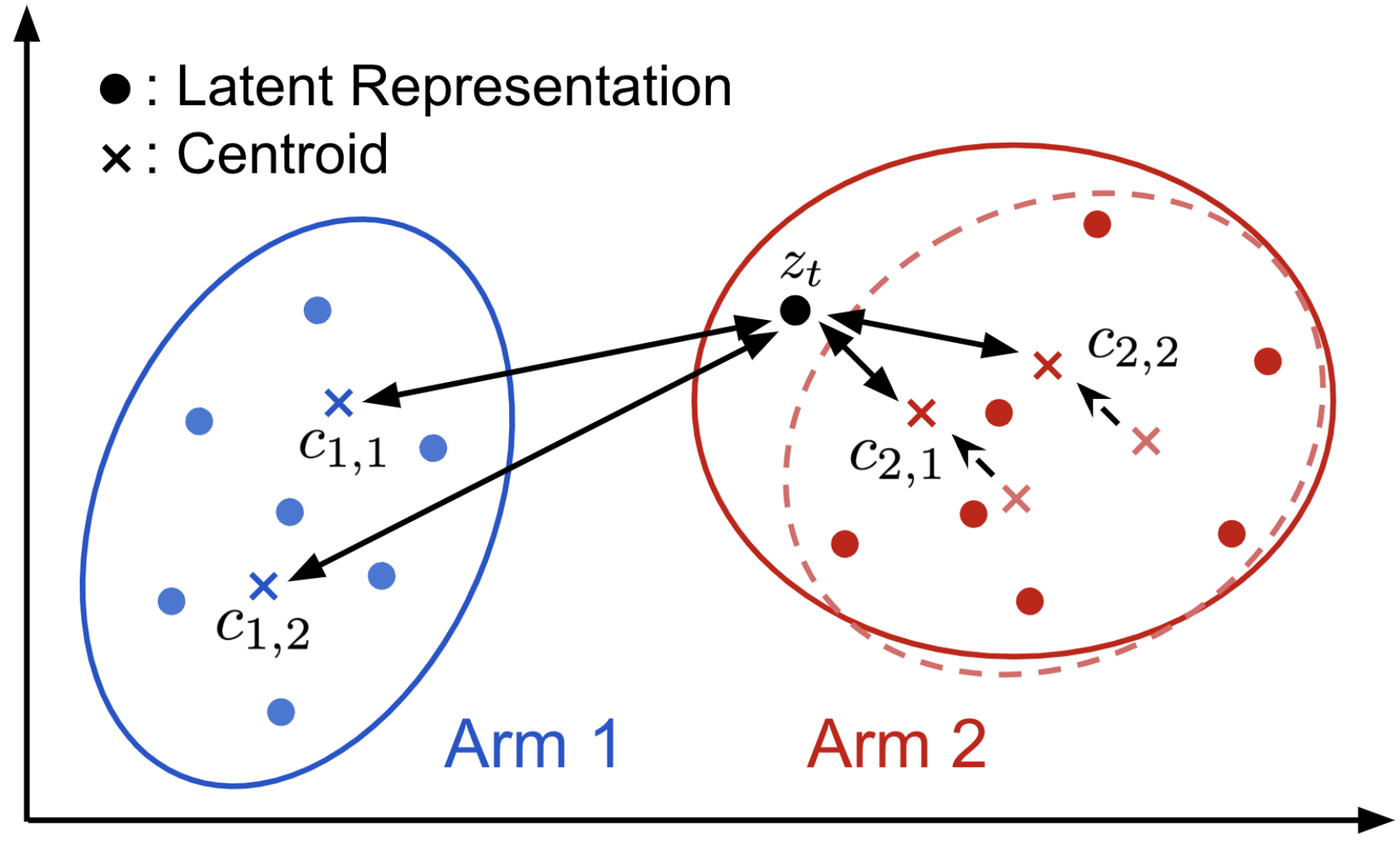}
        \caption{Overview of Centroid Update in k-means.}
    \label{fig:kmeans}
    \end{center}
\end{figure}

%
$\mathrm{NeuralRS}$ assigns multiple centroids to each action and estimates the trial ratio (reliability) by clustering feature vectors using k-means centroid update.
%
The latent representation used for the approximation is taken from the output of the penultimate layer of the NN.
%
By leveraging k-means centorid updates, the centroids can capture the entire feature space, allowing the agent to balance exploration and exploitaion while improving long-term adaptability.
%
Using centroids also helps the agent to adjust to biases or density in the feature space, enabling stable action selection and reducing the risk of getting stuck in local optimal solutions.
%
The trial ratio is approximated by computing weights based on Euclidean distances and applying these weights to the selection counts for each cluster.
%
Specifically, at time step $t$, the Euclidean distance between the latent representation $z_t$ and each centroid $c_{i, m}$ assigned to action $a_i$ is calculated using Eq. \eqref{eq:distance}.
\begin{align}
    \label{eq:distance}
    d_{i, m} &= \| z_t - c_{i, m} \|
\end{align}
%
The weight $w_{i,k}$ is then defined as in Eq. \eqref{eq:weight}.
\begin{align}
    \label{eq:weight}
    w_{i, m} &= \frac{1}{d_{i, m} + \epsilon}
\end{align}
%
where $\epsilon$ is a small constant to avoid division by zero.
%
The approximated trial ratio $\rho_i$ is calculated as the weighted average of the selection counts $n_i$ for each centroid, as shown in Eq. \eqref{eq:rho}.
%
This prevents trials from overly concentrating on specific clusters and is expected to lead to an appropriate trial distribution over time.
\begin{align}
    \label{eq:rho}
    \overline{n}_i &= \frac{n_{i}}{M}  \sum_m w_{i, m} \\
    \rho_i &= {\mathrm{softmax}}(\overline{n}_i)
\end{align}
%
Here, $M$ is the total number of centroids assigned to each action $a_i$.
%
Multiple centroids $c_{i, m}$ are assigned to each action $a_i$.
%
These centroids are initialized according to a normal distribution $c_{i, m} \sim {\mathcal{N}} {(0, {\sigma^2} \mathbf{I})}$, and the total weights $W_{i, m}$ and selection counts $n_i$ are initially set to zero.
%
Based on the latent representation $z_t$ at time $t$, the weights $w_{i, m}$ for each centroid $c_{i, m}$ are calculated. Using these weights, the centroids are updated according to Eq. \eqref{eq:update_centroid}.
\begin{align}
    \label{eq:update_centroid}
    c_{i, m} = \frac{W_{i, m} c_{i, m} + w_{i, m} z_t} {W_{i, m} + w_{i, m}}
\end{align}
%
Figure \ref{fig:kmeans} provides an overview of the centroid updates in k-means.
%
The total weights $W_{i, m}$ and selection counts $n_i$ for each cluster are updated with a time-decay factor $\gamma$, allowing the influence of past data to gradually diminish while adapting to new data.

\section{Reliability candidates}
%
Several approaches can be used to calculate reliability in $\mathrm{NeuralRS}$.
%
Reliability is a key factor in optimizing trial ratios and can influence the agent's decision-making in different ways.
%
In this study, we consider several candidates for reliability calculation, including k-means and kNN introduced in $\mathrm{NeuralRS}$ and $\mathrm{RegLinRS}$, as well as Cross-Entropy and Trial Ratio, which uses the actual counts of selected actions.

\subsection{Reliability Based on Cross-Entropy (XE)}
%
The Cross-Entropy(XE) based method evaluates reliability using the selection probability for each action, as determined by the output of the NN.
%
This approach is expected to provide a measure of the uncertainly of each action based on the model's output, making it useful for assessing the reliability of each action.
%
Reliability based on XE becomes more stable as the model's learning progresses, serving as a reliable foundation for long-term action selection.
%
This method helps prevent excessive exploration and supports efficient action selection, ensuring a balance between exploration and exploition.

\subsection{Reliability Based on Trial Ratio}
%
The method based on Trial Ratio directly uses the trial ratio of each action, which is calculated past trial data, as its reliability.
%
This method adjusts the reliability based on frequency, making it easier for the agent to prioritize actions with sufficient trials.
%
Reliability estimation based on Trial Ratio ensures stable action selection and prevents any action from being unfairly assigned a low reliability.
%
Additionally, the Trial Ratio based reliability enables the agent long-term adaptation, with the reliability evaluation becoming more accurate as the number of trials increases.
%
This mechanism supports effective agent behavior in the long term by ensuring stable action selection and balancing exploration and exploitation efficiently.

\section{Experiments}
%
We evaluated the performance of the proposed method, $\mathrm{NeuralRS}$, using two datasets: the artificial dataset which sampled from an artificially generated distribution and the Statlog-Shuttle dataset from UCI \cite{dua17} as a real-world dataset.

\subsection{Artificial dataset}
%
To compare $\mathrm{NeuralRS}$ and $\mathrm{RegLinRS}$ against benchmark methods using the same evaluation metrics, we created an artificial dataset based on the setup described in \cite{tsuboya24}, where the aspiration level $\aleph$ remains constant.
%
The dataset assumes a linear relationship between the feature vector and the expected rewards of each action. The parameters used were follows: the feature vector ${\bf x}_t$ had a dimension of $64$, the number of actions was $K = 4$, the optimal aspiration level was $\aleph = 0.7$, and the total number of data points was 10,000.
%
At each time step $t$, the agent received a feature ${\bf x}_t$ with a small addition of noise.
%
Based on this feature vector, the agent selected one action from $K = 4$ options and received a reward.
%
%

\subsection{Statlog-Shuttle dataset}
%
The Statlog-Shuttle dataset consists of $10$-dimensional feature vectors collected from space shuttle sensors and is used to classify normal states and abnormal states (e.g., failures or anomalies).
%
Appoximately $80 \%$ of the data belong to class $1$, meaning that simply predicting class 1 yields a default accuracy of about $80 \%$.
%
The objective for the model when using this dataset is to achieve a classification accuracy exceeding $99 \%$, an extremely accurate performance that exceeds the default performance.

\subsection{Experimental settings}
%
The agent performed 10,000 steps, with each step consisting of an action selection and receiving feedback on the reward.
%
We defined it as one simulation, and the average results were calculated over $100$ repeated simulations.
%
For NN-based algorithms, the approximation function used a neural network with 128 units in the hidden layer and $K$ units in the output layer. The network had two fully connected layers. Training was conducted at each step with a batch size of 1,024.
%
We used Adam as the optimizer, with a learning rate of $1 \times 10^{-3}$.
%
For linear appoximation methods, the batch size was also set to $20$.
%
To initialize the parameters of each method, the agent chose each action 10 times immediately after the start of a simulation.
%
The reliability measure used in $\mathrm{NeuralRS}$ was chosen based on the best-performing method.
%
The parameters used in the experiment are shown in Table \ref{tbl:parameter_values}.
\begin{table}[hbtp]
    \caption{Configuration values and initial values.}
    \label{tbl:parameter_values}
    \hspace{0.3cm}
    \begin{tabular}{cc}
        \hline
        Variable & Value \\
        \hline
        \hline
        $\aleph$ & $0.65$ \\
        $M$ & $K \times 2$ \\
        $\gamma$ & $0.99$ \\
        $\nu$ of $\mathrm{NeuralUCB}$ $\And$ $\mathrm{NeuralTS}$ & $0.1$ \\
        $\lambda$ of $\mathrm{NeuralUCB}$ $\And$ $\mathrm{NeuralTS}$ & $1 \times 10^{-5}$ \\
        episodic memory size of $\mathrm{RegLinRS}$ & 10,000 \\
        $k$ of $K$-nearest neighbors of $\mathrm{RegLinRS}$ & $50$ \\
        $\alpha$ of $\mathrm{LinUCB}$ & $0.1$ \\
        $\lambda$ of $\mathrm{LinTS}$ & $0.25$ \\
        $\alpha$ of $\mathrm{LinTS}$ & $6$ \\
        $\beta$ of $\mathrm{LinTS}$ & $6$ \\
        \hline
    \end{tabular}
\end{table}

\subsection{Results}
\begin{figure}[hbtp]
    \begin{center}
        \includegraphics[width=7.5cm]{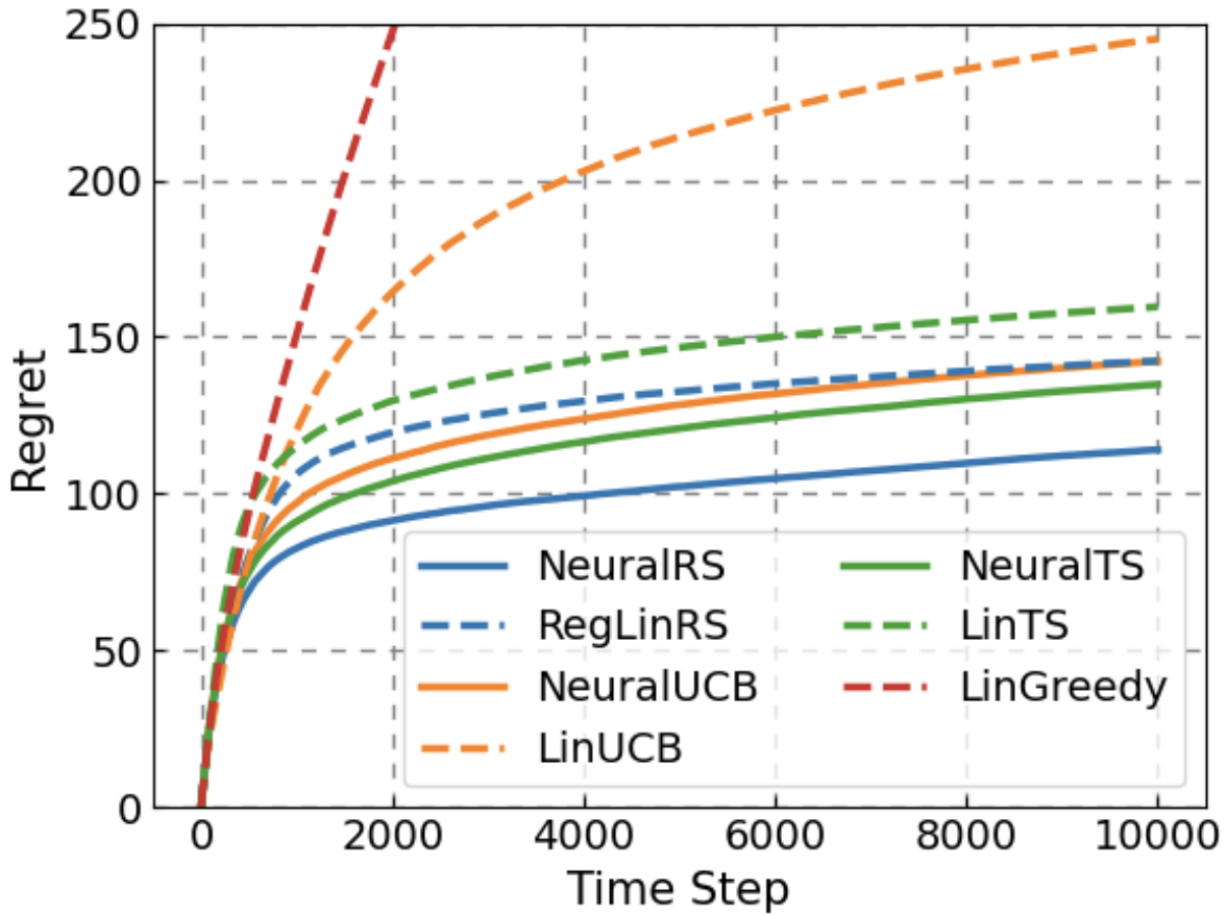}
        \caption{Comparison of NeuralRS and Baseline Algorithms on an Artificial Dataset.}
    \label{fig:ex_regret_artificial}
    \end{center}
\end{figure}

\begin{figure}[hbtp]
    \begin{center}
        \includegraphics[width=7.5cm]{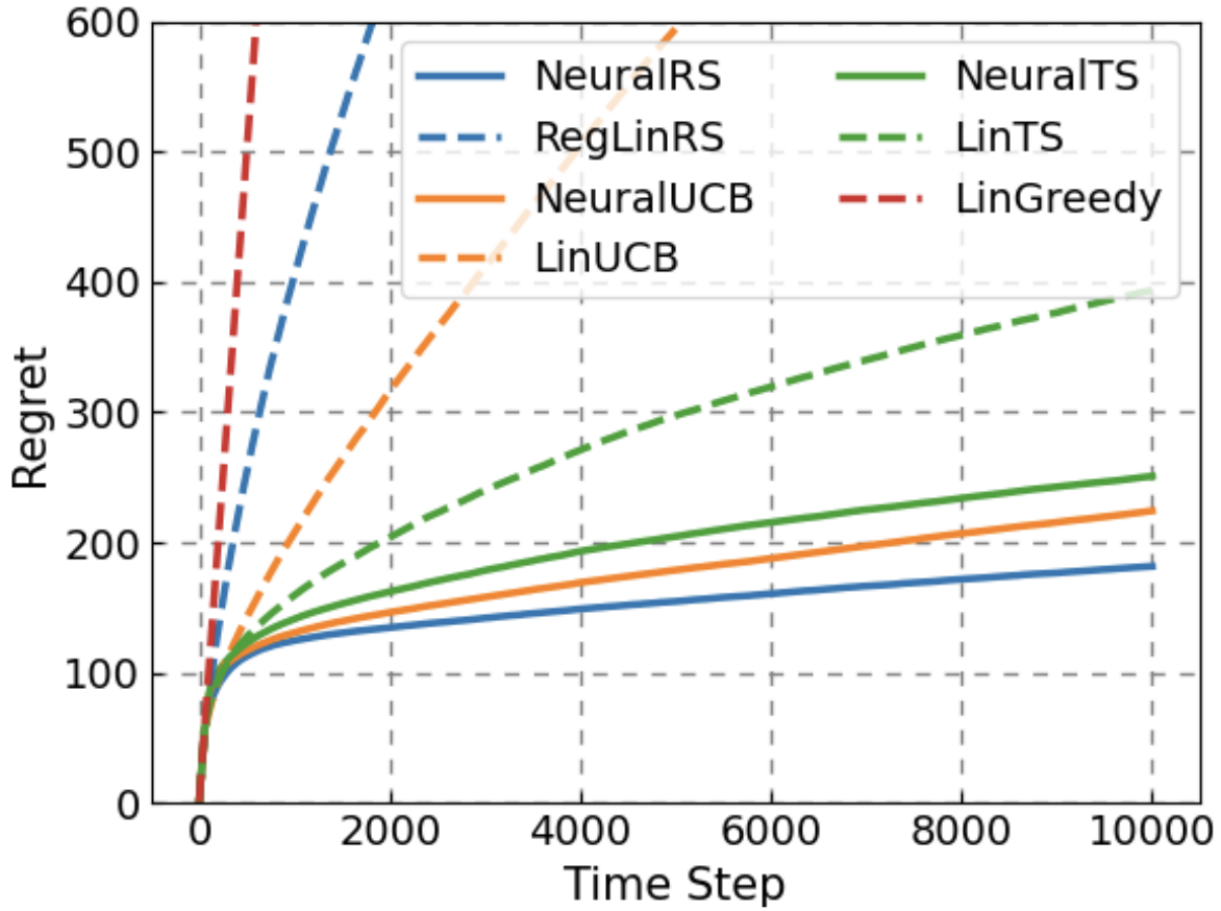}
        \caption{Comparison of NeuralRS and Baseline Algorithms on the Statlog-Shuttle Dataset.}
    \label{fig:ex_regret_shuttle}
    \end{center}
\end{figure}

%
The trends of regret for each method on the artificial dataset and the Statlog-Shuttle dataset are shown in Figure \ref{fig:ex_regret_artificial} and Figure \ref{fig:ex_regret_shuttle}, respectively.
%
Figures \ref{fig:ex_regret_artificial} and \ref{fig:ex_regret_shuttle} show that $\mathrm{NeuralRS}$ demonstrates the lowest regret compared to all other benchmark algorithms.
%
Neural-based algorithms outperform their corresponding linear approximation algorithms.
%
This result suggests that incorporating NN enables the model to learn more complex state representations and to estimate expected rewards more accurately than linear approximations.
%
Especially in Figure \ref{fig:ex_regret_shuttle}, $\mathrm{NeuralRS}$ shows significantly improved performance compared to $\mathrm{RegLinRS}$.
%
As we mentioned in the previous section, there are multiple approaches for estimating reliability used in $\mathrm{NeuralRS}$.
%
In the next section, we will discuss the comparison of different reliability estimation methods for $\mathrm{NeuralRS}$ using the artificial dataset and Statlog-Shuttle dataset.

\section{Discussion}
\begin{figure}[hbtp]
    \begin{center}
        \includegraphics[width=7.5cm]{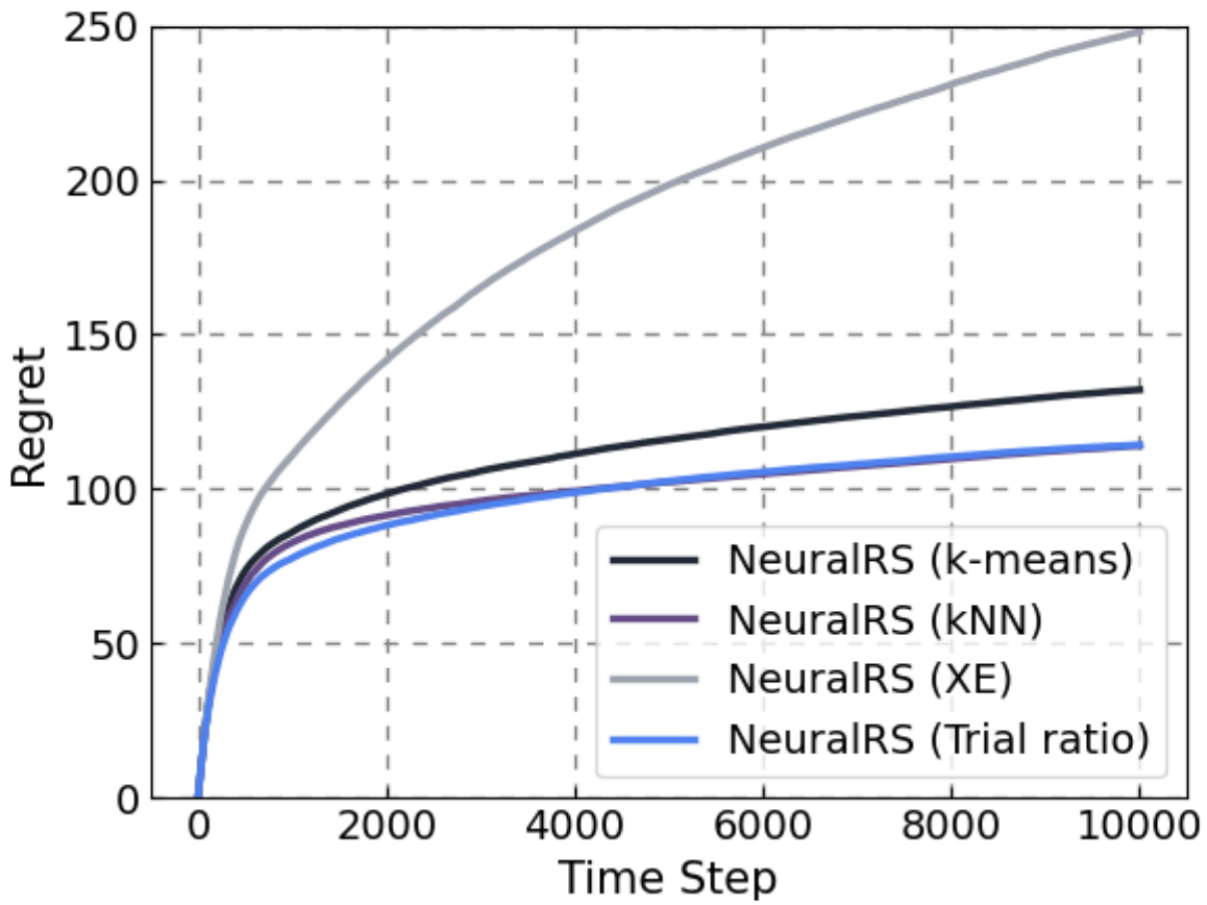}
        \caption{Comparison of Reliability Candidates for NeuralRS on an Artificial Dataset.}
    \label{fig:dis_regret_artificial}
    \end{center}
\end{figure}

\begin{figure}[hbtp]
    \begin{center}
        \includegraphics[width=7.5cm]{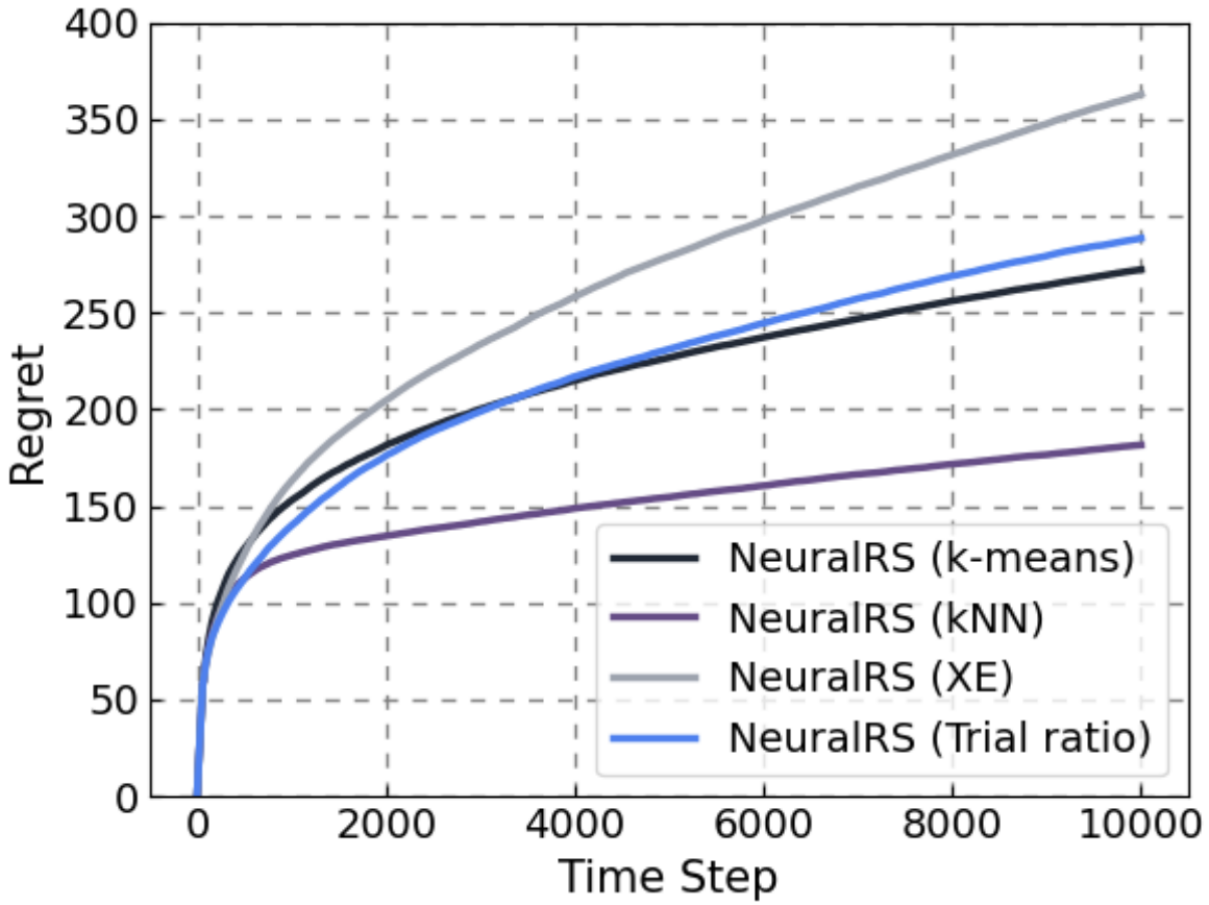}
        \caption{Comparison of Reliability Candidates for NeuralRS on the Statlog-Shuttle Dataset.}
    \label{fig:dis_regret_shuttle}
    \end{center}
\end{figure}

%
The regret trends for each reliability candidate in $\mathrm{NeuralRS}$ on the artificial dataset and Statlog-Shuttle dataset are shown in Figure \ref{fig:dis_regret_artificial} and \ref{fig:dis_regret_shuttle}, respectively.
%
For both datasets, reliability estimation that used kNN showed the best performance.
%
In this experiment, k-means resulted in higher regret compared to kNN.
%
However, centroids obtained through k-means could capture the overall trends in the feature space, focusing more on global tendencies than on local fluctuations.
%
This makes k-means-based reliablity estimation potentially more effective, relative to other methods, in noisy environments or with large datasets.
%
Additionally, since k-means used fixed centroids, it was less computationally demanding compared to kNN, making it more efficient for large scale data processing.
%
This indicates that k-means based reliablity estimation is particularly suitable for systems that reaure real-time performance or for environments with limited computional resources.
%
From Figure \ref{fig:dis_regret_artificial}, it can be observed that Trial Ratio, which has less fluctuation in reliablity, also performed well.
%
This is likely because the artificial dataset had a relatively simple structure, which allowed action selection based on expected rewards to function effectively.
%
Although action selection based on expected rewards was effective, $\mathrm{LinGreedy}$ showed increased regret. This was likely due to small differences in reliablity that had a significant impact on minimizing regret.
%
In contrast, as shown in Figure \ref{fig:dis_regret_shuttle}, in a real-world dataset like the Statlog-Shuttle dataset, reliablity based on Trial Ratio performed worse compared to the other methods, particularly kNN and k-means.
%
These results suggest that complex environments require more careful computation of reliablity during action selection.
%
From Fugure \ref{fig:dis_regret_artificial} and \ref{fig:dis_regret_shuttle}, reliability using XE showed the lowest performance.
%
This is likely because XE-based reliablity depends on NN outputs, which can be unstable in the earlier time steps when sample sizes are small, leading to biased evaluations for certain actions or states.

\section{Conclusion}
%
We proposed $\mathrm{NeuralRS}$, which incorporates Neural Network (NN) fuction approximation into the previously proposed $\mathrm{RegLinRS}$ algorithm for contextual bandit problems.
%
By introducing NN, $\mathrm{NeuralRS}$ demostrated the ability to handle more diverse environments compared to the linear approaches of $\mathrm{RegLinRS}$.
%
$\mathrm{NeuralRS}$ also showed a capability to quickly switch to exploitation while minimizing exploration. It adapted to environments more flexibly than other NN-based algorithms such as $\mathrm{NeuralUCB}$ and $\mathrm{NeuralTS}$.
%
These results suggest that $\mathrm{NeuralRS}$ is particularly useful for real-world applications where the cost of exploration is high.
%
We also examined the impact of different reliablity estimation methods on the performance.
%
Although k-means-based reliability performed worse than kNN-based methods in the short term, k-means centroids capture overall trends in the feature space more effectively. This makes it potentially advantageous in noisy environments or with large datasets.
%
Moreover, the low computational cost of k-means-based reliability estimation makes it suitable for systems that require real-time processing or operating in environments with limited computational resources.
%
Our future works will include evaluating k-means-based reliability estimation on large, noisy, and complex datasets to confirm its practicality and pursuing real-world applications.



\begin{thebibliography}{9}

\bibitem{takahashi16}
T.~Takahashi, Y.~Kono, and D.~Uragami (2016) ``Cognitive satisficing: bounded
  rationality in reinforcement learning,'' \textit{Transactions of the Japanese
  Society for Artificial Intelligence}, 31 (6),  AI30--M\_1, Num Pages: 11 (in Japanese).

\bibitem{tsuboya24}
A.~Tsuboya, Y.~Kono, and T.~Takahashi (2024) ``A {Sequential} {Decision}-{Making} {Model} in {Contextual} {Foraging} {Behavior},'' \textit{journal of Japan Society for Fuzzy Theory and Intelligent Informatics}, 36 (1), 589--600, Num Pages: 12 (in Japanese).

\bibitem{li10}
L.~Li, W.~Chu, J.~Langford, and R.~E. Schapire (2010) ``A contextual-bandit
  approach to personalized news article recommendation,'' in
  \textit{Proceedings of the 19th international conference on {World} wide
  web}, {WWW} '10,  661--670, New York, NY, USA: Association for Computing
  Machinery, April.

\bibitem{riquelme18}
C.~Riquelme, G.~Tucker, and J.~Snoek (2018) ``Deep {Bayesian} {Bandits}
{Showdown}: {An} {Empirical} {Comparison} of {Bayesian} {Deep} {Networks} for
  {Thompson} {Sampling},'' in  \textit{International Conference on Learning
  Representations}.

\bibitem{zhou20}
D.~Zhou, L.~Li, and Q.~Gu (2020) ``Neural {Contextual} {Bandits} with {UCB}-based {Exploration},'' in \textit{Proceedings of the 37th International Conference on Machine Learning}, {ICML} '20,  11492--11502, Virtual: Proceedings of Machine Learning Research, July.

\bibitem{zhang21}
T.~Zhang, C.~Yang, and Q.~Gu (2021) ``Neural {Thompson} {Sampling},'' in \textit{Proceedings of the International Conference on Learning Representations}, {ICLR} '21, Virtual, April.

\bibitem{simon56}
H.~A. Simon.
\newblock Rational choice and the structure of the environment.
\newblock {\em Psychological Review}, 63:129--138, 1956.

\bibitem{tamatsukuri19}
A.~Tamatsukuri and T.~Takahashi.
\newblock Guaranteed satisficing and finite regret: Analysis of a cognitive
  satisficing value function.
\newblock {\em Biosystems}, 180:46--53, 2019.

\bibitem{dua17}
D.~Dua and C.~Graff (2017) ``{UCI Machine Learning Repository},'' in \textit{University of California, Irvine}, {ICS}.

\end{thebibliography}
\end{document}